\documentclass{fairmeta}
\usepackage{amsmath}
\usepackage{enumerate} 
\usepackage{bm}
\usepackage{algorithm}
\usepackage{floatflt}
\usepackage{algpseudocode}
\usepackage{amsfonts}
\usepackage{amsthm}
\usepackage{newtxtt}
\usepackage{algorithm}
\usepackage{colortbl} 
\usepackage{cleveref}
\usepackage{diagbox} 
\usepackage[utf8]{inputenc}
\usepackage{textgreek}
\usepackage{colortbl}
\usepackage{nicematrix}
\usepackage{makecell}
\usepackage{float}
\usepackage{arydshln}
\usepackage[frozencache,cachedir=.]{minted}
\usepackage{caption}
\usepackage{subcaption}
\usepackage{tcolorbox}
\usepackage{amssymb}
\usepackage{xspace}
\usepackage{wrapfig}
\usepackage{adjustbox}
\usepackage{tabularx}
\usepackage{booktabs}
\usepackage{mathtools}
\usepackage{wrapfig}
\usepackage{amssymb}
\usepackage{graphicx}

\usepackage{silence}
\makeatletter
\patchcmd{\wrong@fontshape}{\@gobbletwo}{}{}{}
\makeatother
\definecolor{upColor}{RGB}{17,138,21}
\definecolor{downColor}{RGB}{174,36,67}

\newtheorem{theorem}{Theorem}[]

\newtheorem{remark1}[theorem]{Remark}

\title{Information Capacity: Evaluating the Efficiency of Large Language Models via Text Compression}
\author[]{Cheng Yuan}
\author[]{Jiawei Shao}
\author[]{Xuelong Li}
\affiliation[]{Institute of Artificial Intelligence (TeleAI), China Telecom}
\code{\url{https://github.com/TeleAI-AI-Flow/InformationCapacity}}
\dataset{\url{https://huggingface.co/datasets/TeleAI-AI-Flow/InformationCapacity}}
\metadata[Correspondence to]{Xuelong Li (\email{xuelong\_li@ieee.org})}

\begin{document}

\abstract{
Recent years have witnessed the rapid advancements of large language models (LLMs) and their expanding applications, leading to soaring demands for computational resources. The widespread adoption of test-time scaling further intensifies the tension between model capability and resource consumption. However, a rigorous metric that accurately reflects an LLM's inference efficiency across diverse tokenizers, parameter counts, and model architectures remains absent. Motivated by the correlation between compression and intelligence, we introduce \textbf{information capacity}, a measure of model efficiency based on text compression performance relative to computational complexity. A distinctive feature of information capacity is its incorporation of tokenizer efficiency, which affects inference costs but is often neglected in LLM evaluations. We assess the information capacity of 56 open-source models and observe a consistent information capacity among different-sized models within a series. Experiments on five heterogeneous datasets reveal strong linguistic biases in mainstream LLMs. Empirical results verify the accuracy of performance prediction across model sizes based on information capacity and show the correlation between information capacity and benchmark scores. This metric can be used to quantify improvements in inference efficiency and provide insights into better scaling performance for future LLM development.
}

\maketitle

\section{Introduction}

The recent advancements in large language models (LLMs) give birth to sophisticated capabilities in reasoning, coding, and tool use, propelling their widespread adoption in various downstream tasks~\citep{huynh2025largelanguagemodelscode,luo2025largelanguagemodelagent,Vicinagearth_llm_tool_use_survey,wang2025aiagenticprogrammingsurvey}.
To handle soaring inference requests, giant computing clusters are being built at an unprecedented speed, which incurs enormous energy consumption that poses significant environmental and economic challenges~\citep{energy_and_AI,IEA_AI}.
Moreover, inference-time scaling~\citep{snell2024scalingllmtesttimecompute} has proven necessary for the advanced thinking and agentic capabilities of the latest LLMs, such as long-horizon planning~\citep{belle2025agentschangeselfevolvingllm,Vicinagearth_llm_mas_survey} and autonomous coding~\citep{yang2025codethinkthinkcode,anthropic2025claudesonnet45}. 
These breakthroughs significantly extend the input and output lengths of LLMs and intensify the tension between model capability and computational costs~\citep{sardana2025chinchillaoptimalaccountinginferencelanguage}.
Additionally, the excessive inference delay leads to insufficient responsiveness, which hinders delay-critical tasks and degrades user experience~\citep{liang2025largemodelempoweredembodied,embodiedai_survey}.
To address these bottlenecks, major corporations have devoted increasing efforts to developing efficient LLMs with strong capabilities but low computational costs~\citep{anthropic2025haiku45,minimax2025minimaxm2}.
However, this field lacks a rigorous method to evaluate model efficiency across diverse tokenizers, parameter counts, and model architectures. 
Current metrics~\citep{xiao2024densinglawllms} fail to bridge the gap between parameter count and inference cost due to differences in both network structure and tokenizer design.

Motivated by \textit{the correlation between compression and intelligence}~\citep{delétang2024languagemodelingcompression}, we evaluate an LLM's efficiency from the perspective of compression.
Mainstream LLMs utilize the decoder-only transformer architecture, which predicts the probability distribution of the next token given the preceding context and delivers remarkable performance in generative and understanding tasks~\citep{vaswani2017attention,brown2020languagemodelsfewshotlearners}.
On the other hand, probability prediction is also the cornerstone of lossless compression.
Modern entropy coding methods, such as arithmetic coding~\citep{langdon1984arithmeticcoding} and asymmetric numeral system (ANS)~\citep{duda2009asymmetricnumeralsystems}, can reduce the number of bits required for representing the given data to its negative $\log_2$-likelihood, optimal as per Shannon's source coding theorem~\citep{shannon1948communicationtheory}.
Consequently, minimizing the encoded symbol length is equivalent to maximizing the predicted likelihood, and the efficiency of lossless compression is determined by how well the probabilistic model can predict subsequent data~\citep{delétang2024languagemodelingcompression}.

Equipped with strong probability prediction capabilities, LLMs are deemed suitable for the probabilistic model of entropy coding~\citep{chen2025informationcompression}.
The pretraining of LLMs aims to minimize the cross-entropy loss on plain text, thus enhancing the model's ability to predict the next token.
This objective can also be viewed as minimizing the negative $\log_2$-likelihood, i.e., the symbol length after entropy coding, on the pretraining corpus~\citep{pan2025understandingllmbehaviorscompression}.
The direct linkage between text compression and pretraining loss suggests a strong correlation between compression and intelligence, which is empirically validated through quantitative measurements~\citep{huang2024compressionrepresentsintelligencelinearly}.
Existing studies have demonstrated the performance advantage of using LLMs to compress text~\citep{valmeekam2023llmziplosslesstextcompression,narashiman2024alphazipneuralnetworkenhancedlossless,mittu2024finezippushinglimits}, audio~\citep{li2025losslessdatacompression}, images~\citep{delétang2024languagemodelingcompression}, and a mixture of these sources~\citep{heurteldepeiges2025compressionpretrainedtransformersstudy}.

\begin{figure*}[!t]
    \centering
    \includegraphics[width=\linewidth]{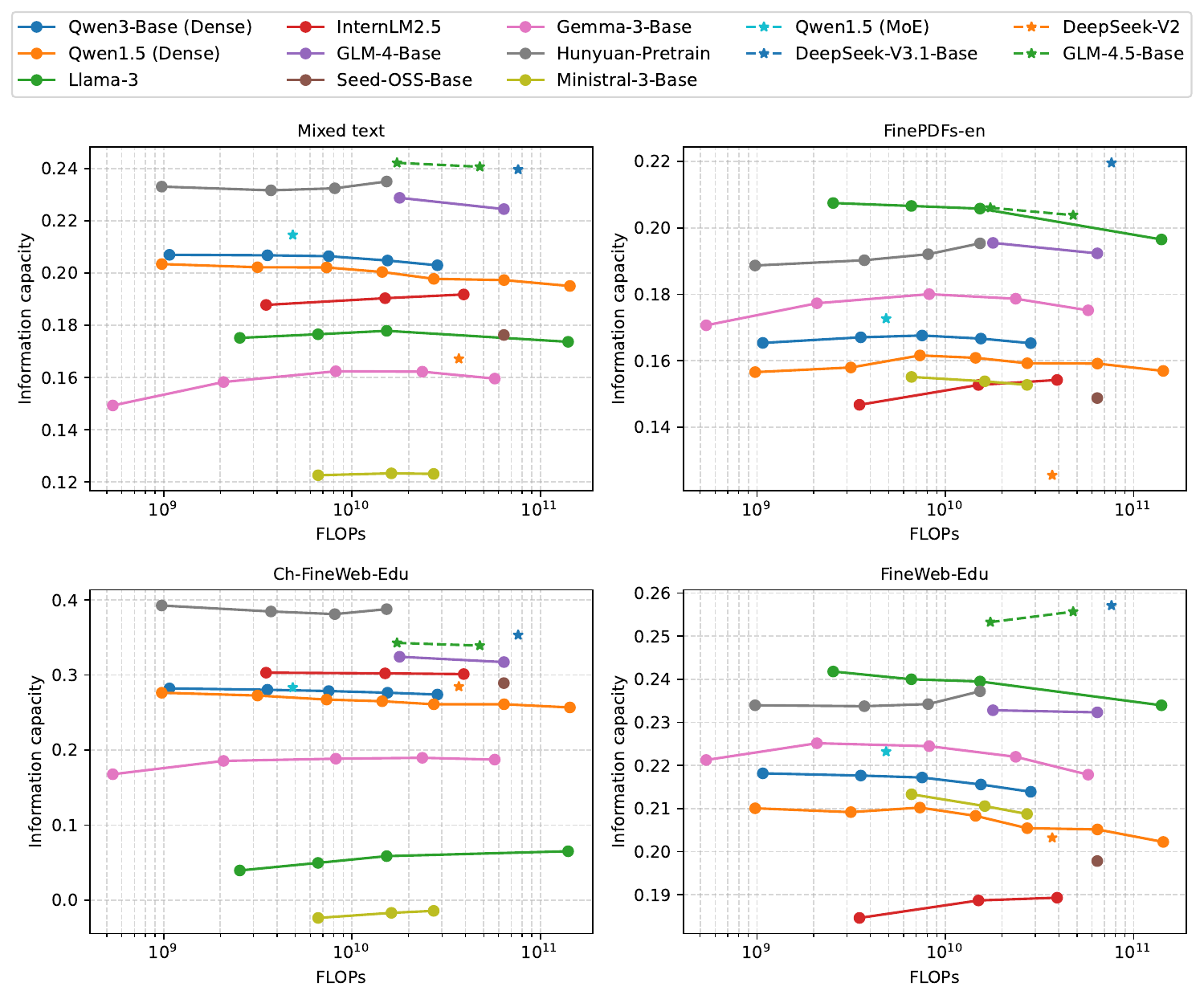}
    \caption{\textbf{Information capacity of mainstream open-source models.} Motivated by the strong correlation between compression and intelligence, information capacity evaluates an LLM's efficiency by text compression performance relative to its computational complexity. Larger models can predict the next token more accurately, leading to higher compression gains but at increased computational costs. Consequently, a series of models with varying sizes exhibits consistent information capacity, which can be used to compare model capability across model series and predict model performance within a series.}
    \label{fig:ic_mixed}
\end{figure*}

In this paper, we introduce \textbf{information capacity}, which evaluates an LLM's \textit{efficiency} based on text compression performance relative to its computational complexity.
Larger models can predict the next token more accurately, leading to greater compression gains but at higher computational costs. 
Consequently, a series of models with varying sizes exhibits consistent information capacity, as shown in Figure~\ref{fig:ic_mixed}.
Information capacity can be used to compare inference efficiency across model series and predict model performance within a series.
Compared to existing metrics such as capability density~\citep{xiao2024densinglawllms}, a key difference of information capacity is that it considers the influence of \textit{tokenizer efficiency}.
An effective tokenizer can represent a given text with fewer tokens, thus reducing both the input and output token counts.
This reduction not only lowers computational costs and inference delay but also facilitates long-context memory and in-depth reasoning~\citep{levy-etal-2024-task}.
In light of the exploding input length and the widespread usage of test-time scaling~\citep{snell2024scalingllmtesttimecompute}, tokenizer efficiency exhibits growing significance but is often neglected in LLM evaluations.

We assess the information capacity of 56 models across 5 heterogeneous datasets, which reveals strong linguistic biases in mainstream LLMs.
Consistent influences on information capacity are observed with respect to tokenizer efficiency, pretraining data, and the mixture-of-experts (MoE) architecture.
Empirical results verify the accuracy of performance prediction across model sizes based on consistent information capacity and show the correlation between information capacity and benchmark scores.
We also conduct ablation studies on the impacts of post-training, test sample length, and softmax temperature, respectively.
Information capacity serves as a unified metric of inference efficiency across model sizes and architectures (e.g., dense and MoE models), which enables cross-scale comparison for LLM evaluation.
This feature exhibits growing relevance, as intelligent services are increasingly deployed on heterogeneous hardware, such as the three-tier network architecture studied in the \textbf{AI Flow} framework~\citep{aiflow, aiflow_perspectives}, to address privacy and latency concerns~\citep{qu2025mobileedgeintelligencelarge, luo2025edgegeneralintelligencemultiplelarge}.
This metric can be used to quantify improvements in inference efficiency and provide insights into better scaling performance for future LLM development.

\section{Preliminaries}
\subsection{Lossless Compression}

For a stream of data $x_{1:L} := x_1 x_2 \ldots x_L$ with length $L$ from a finite symbol set $\mathcal{X}$, the probability of the whole sequence $x_{1:L}$ is the product of the conditional probabilities of all constituting symbols given previous data, written as $p(x_{1:L}) = \prod_{i=1}^{L} p(x_i | x_{<i})$.
Entropy coding reduces the size of input data by assigning a shorter binary codeword to a more frequent sequence, and the latest methods produce a bitstream with a length approximating the negative $\log_2$-likelihood of a given distribution $p(x)$, expressed as $- \log_2 p(x_{1:L}) = \sum_{i=1}^{L} -\log_2 p(x_i | x_{<i})$.
In reality, the coding algorithm introduces marginal overhead. 
The actual code length for arithmetic coding is $(-\text{roundup}(\log_2 p(x_{1:L}))+1)$ bits to avoid boundary ambiguity \citep{langdon1984arithmeticcoding}, and the $B$-bit precision calculation of arithmetic operations involved further adds $O(2^{-B}L)$ bits overhead \citep{delétang2024languagemodelingcompression}. 
On the other hand, ANS uses the discretized probabilities derived from the integer frequencies of input symbols $\bar{p}(x)$ for entropy coding, where the discretization errors cause a small overhead in bit length \citep{duda2009asymmetricnumeralsystems}.

Note that the true source distribution $\rho(x)$ is untractable and can only be approximated by a probability model.
The more accurate is the estimated probability $p(x)$, the closer is the encoded symbol length $-\log_2 p(x_{1:L})$ to the theoretical optimum $-\log_2 \rho(x_{1:L})$ \citep{shannon1948communicationtheory}.
Additionally, only previous symbols have been decoded and are available to the probability model during the sequential decoding process.
This causality essentially requires the probability model to iteratively predict the next symbol given previous symbols, which aligns well with decoder-only LLMs.

\subsection{LLMs as Probability Estimators} \label{sec:llm_probability_estimator}

Mainstream LLMs employ the decoder-only transformer architecture, consisting of an embedding layer, several transformer blocks, and a language model head.
For the $i$-th token $x_i$ in the input sequence of length $L$ ($i = 1, 2, \ldots, L$), the LLM $M$ outputs a vector called logits, whose length is equal to the vocabulary size\footnote{In reality, the length of logits is slightly larger than the tokenizer's vocabulary size. Check Section~\ref{sec:evaluation_method} for details.}.
A softmax function converts logits into estimated probabilities of the next token for all possible tokens in the vocabulary.
Decoder-only LLMs enforce a causal mask on the attention calculation in each transformer block, and this autoregressive prediction maintains causality during decoding. 
Consequently, the estimated probabilities are conditioned on all previous tokens $x_{<i}$, and the estimated probability of the true $i$-th token in the input text is denoted as $p(x_i | x_{<i} ; M)$.

The pretraining of LLMs maximizes the estimated likelihood of the next token by optimizing the cross-entropy loss, expressed as:
\begin{equation}
    H(\rho_{\text{data}}, p) = \mathbb{E}_{x \sim \rho_{\text{data}}}[\sum_{i=1}^{n} -\log_2 p(x_i | x_{<i})] ,
    \label{eq:cross_entropy}
\end{equation}
where $\rho_{\text{data}}$ denotes the probability distribution of the pretraining dataset, and $n$ represents the token length of text samples in the pretraining dataset.
From the perspective of lossless compression, the cross-entropy in (\ref{eq:cross_entropy}) is equal to the expected length of the encoded bitstream on the pretraining data distribution\footnote{The default implementation of deep learning libraries commonly utilizes a natural logarithm function rather than a base of 2, which only applies a constant multiplication on the loss value and does not affect the equivalence in training objective.}.
As a result, the pretraining of LLMs also minimizes the data size when the estimated probabilities are used for entropy coding.

To compress multimodal data with decoder-only LLMs, there are mainly two approaches.
The first approach is to represent the raw bits of multimodal data by text and directly process such pseudo-text with LLMs.
For example, a segment of 7 bits can be mapped to a valid ASCII character that an LLM can recognize \citep{delétang2024languagemodelingcompression}, and any discrete representation of multimodal data can be converted in this manner.
Quantitative evaluations have demonstrated that this approach achieves superior compression rate in images and audio, compared to mainstream lossless compressors, PNG and FLAC, in their respective domains.
The second approach is to directly train decoder-only transformers on datasets in the respective domain, exemplified by the image generative pre-trained transformer (GPT) trained on RGB images \citep{iGPT} and the byte GPT trained on byte sequences of multimodal data \citep{wu2024languagemodelsbytemodels}.
This approach has proven advantageous against both traditional codecs and the first approach \citep{li2025losslessdatacompression}.
Large multimodal models (LMMs) that use a discrete representation of input data for multimodal understanding tasks \citep{cui2025emu35nativemultimodalmodels} are also promising candidates for compressing data in modalities other than text.

\section{Information Capacity}
\subsection{Computation Formula}

Motivated by the strong correlation between compression and intelligence, information capacity evaluates an LLM's efficiency via text compression, defined as the ratio of model intelligence to the inference complexity, given by:
\begin{equation}
    \text{Information Capacity} = \frac{\text{Model Intelligence}}{\text{Model Inference Complexity}} .
    \label{eq:ic_rationale}
\end{equation}
Specifically, the model intelligence is measured by the data size savings achieved from the LLM's probability prediction.
The original size of a text sample in the given dataset is denoted as $C$, which is transformed into a sequence of $L$ tokens by the tokenizer of an LLM $M$.
The LLM predicts the probability distribution of the next token given all previous tokens $x_{<i}$ as context, which is used for the arithmetic coding of the actual next token.
The symbol length of the $i$-th token derived from entropy coding is approximately $-\log_2 p(x_i | x_{<i} ; M)$, and the compression gain is the difference between the original data size and the summed symbol length of all tokens.
Additionally, we measure the computational complexity by the inference floating-point operations (FLOPs) $N_M$ on a logarithmic scale, as previous studies find that inference costs are decreasing exponentially for LLMs with equivalent downstream performance \citep{xiao2024densinglawllms}.
In summary, the information capacity is defined as:
\begin{equation}
    \text{Information Capacity} = \frac{C - \sum_{i} -\log_2 p(x_i | x_{<i} ; M)}{ \log_2 N_M} .
    \label{eq:IC}
\end{equation}

In practical measurements, the inference FLOPs $N_M$ and the compression gain are summed for all tokens in the text sample and are approximately proportional to the text sample length $L$.
The denominator $\log_2 N_M$ is FLOPs in a logarithmic scale, while the numerator is the compression gain itself, rendering the value of information capacity calculated from (\ref{eq:IC}) heavily affected by the sample length $L$.
Thus, the original size, negative log-likelihood (NLL), and inference FLOPs are all averaged by token count to exclude the influence of sample length.
In addition, we exclude the first token during measurements, as mainstream decoder-only LLMs can only predict the next token with at least one effective preceding token as context.
The practical computation formula of information capacity is expressed as:
\begin{equation}
    \text{IC}' = \frac{\frac{1}{L-1} (C - \sum_{i=2}^{L} -\log_2 p(x_i | x_{<i} ; M))}{ \log_2 (N_M / (L-1))} .
    \label{eq:IC_compute}
\end{equation}

\begin{figure}[htb]
    \centering
    \vspace{-6pt}
    \includegraphics[width=.8\linewidth]{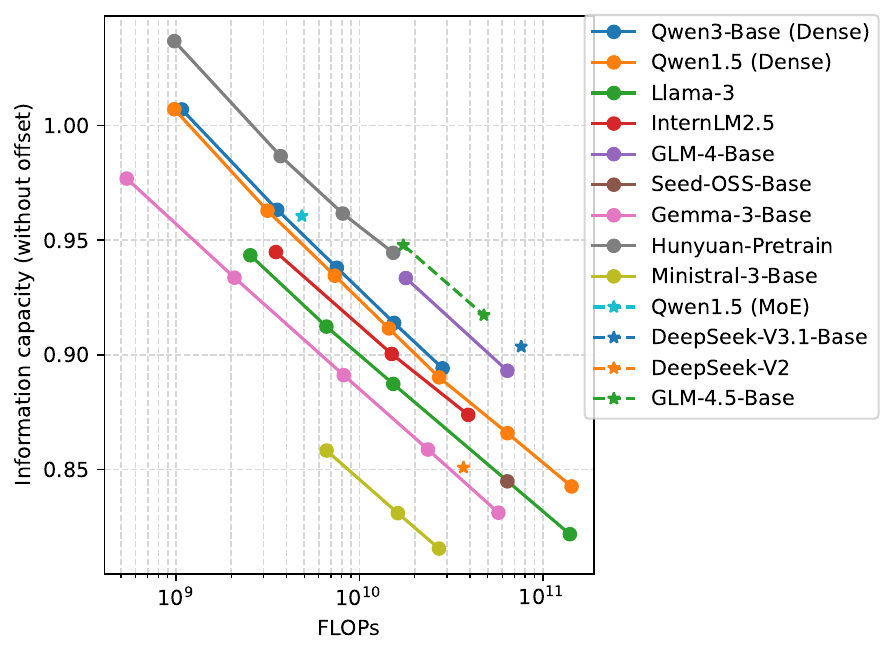}
    \vspace{-6pt}
    \caption{\textbf{Information capacity evaluated on mixed text without numerator offset.} The information capacity calculated from (\ref{eq:IC_compute}) is decreasing almost linearly as the inference FLOPs increase, requiring at least two models to be trained to predict the performance of a different-sized model. Moreover, it is inconvenient to compare model capabilities across different model series.}
    \label{fig:ic_mixed_no_bias}
\end{figure}

As shown in Figure~\ref{fig:ic_mixed_no_bias}, the information capacity calculated from (\ref{eq:IC_compute}) for a model series is decreasing almost linearly as the inference FLOPs increase.
Consequently, two parameters (slope and bias) need to be determined from the measured results, requiring at least two models to be trained to predict the performance of a different-sized model.
Additionally, this linear fitting renders it inconvenient to compare model capabilities across different model series.
To address these limitations, we introduce a negative offset $b$ in the numerator so that different-sized models in a series have nearly identical information capacities, given by:
\begin{equation}
    \text{IC} = \frac{\frac{1}{L-1} (C - \sum_{i=2}^{L} -\log_2 p(x_i | x_{<i} ; M))+b}{ \log_2 (N_M / (L-1))} .
    \label{eq:IC_biased}
\end{equation}
We find that a fixed offset is sufficient to maintain an almost constant information capacity for the influential model series under study, without changing the rankings of different model series.

\subsection{Evaluation Method} \label{sec:evaluation_method}

For each text sample, we measure the original data size $C$ by the average text size per token using the universal UTF-8 encoding.
To maximize evaluation efficiency, we truncate text samples to a fixed sequence length $L = 1024$ to avoid inconsistent token lengths across different models and text samples.
Note that the length of the logits output by modern LLMs is larger than the vocabulary size of the tokenizer, in order to allow the addition of extra tokens and enhance computational efficiency.
These ineffective logits should be truncated before the softmax function to obtain accurate NLLs.
Additionally, we adopt the default temperature setting of $T = 1$ for all evaluated models to ensure a fair comparison. 
Most base models do not specify a temperature value in their \texttt{generation\_config.json} files, and thus use the default value.
For the few models that include a recommended temperature, this value is tuned for the performance of generating new tokens rather than for the probability estimation of given input text.
To enhance numerical precision in calculating the NLLs, we promote the output logits from the original data type \texttt{bfloat16}
to \texttt{float32} with higher precision.

The inference FLOPs are calculated from the hyperparameters based on the model architecture.
We consider an LLM with $l$ transformer blocks, and the dimensions of hidden states, key-value cache, intermediate features in feedforward networks (FFN), and output logits are given by $d_{\text{hid}}$, $d_{\text{kv}}$, $d_{\text{ff}}$, and $d_{\text{logits}}$, respectively.
For MoE models, $d_{\text{ff}}$ is defined as the sum of dimensions for all activated experts.
The total inference FLOPs for mainstream LLMs with grouped-query attention (GQA) \citep{ainslie2023gqatraininggeneralizedmultiquery} and gated FFN are given by:
\begin{equation}
    N_{M} = l (\underbrace{4L d_{\text{hid}} (d_{\text{hid}} + d_{\text{kv}})}_{\text{projection layers}} + \underbrace{4 d_{\text{hid}} \frac{L(L-1)}{2}}_{\text{attention}} + \underbrace{6Ld_{\text{hid}} d_{\text{ff}}}_{\text{FFN}}) + \underbrace{2L d_{\text{hid}} d_{\text{logits}}}_{\text{LM head}} .
    \label{eq:flops}
\end{equation}
For Llama-4 models that introduce auxiliary FFNs for attention scores whose intermediate features have a dimension of $d_{\text{aux}}$ \citep{meta2025llama4}, the FLOPs are calculated as:
\begin{equation}
    N_{\text{Llama-4}} = l (\underbrace{4L d_{\text{hid}} (d_{\text{hid}} + d_{\text{kv}})}_{\text{projection layers}} + \underbrace{4 d_{\text{hid}} \frac{L(L-1)}{2}}_{\text{attention}} + \underbrace{4L d_{\text{hid}} d_{\text{aux}}}_{\text{Auxiliary FFN}} + \underbrace{6Ld_{\text{hid}} d_{\text{ff}}}_{\text{FFN}}) + \underbrace{2L d_{\text{hid}} d_{\text{logits}}}_{\text{LM head}} .
    \label{eq:flops_llama4}
\end{equation}
For DeepSeek models that replace GQA by multi-head latent attention (MLA) whose key-value latents have a dimension of $d_{\text{latent}}$ \citep{deepseekai2024deepseekv2strongeconomicalefficient}, the FLOPs are written as:
\begin{equation}
    N_{\text{DeepSeek}} = l (\underbrace{4L d_{\text{hid}} (d_{\text{hid}} + d_{\text{kv}})}_{\text{projection layers}} + \underbrace{4L d_{\text{hid}} d_{\text{latent}}}_{\text{attention}} + \underbrace{6Ld_{\text{hid}} d_{\text{ff}}}_{\text{FFN}}) + \underbrace{2L d_{\text{hid}} d_{\text{logits}}}_{\text{LM head}} .
    \label{eq:flops_deepseek}
\end{equation}

We comprehensively evaluate the information capacity of different models on five heterogeneous datasets: Mixed text, FinePDFs-en \citep{finepdfs}, Ch-FineWeb-Edu \citep{fineweb-edu-ch}, FineWeb-Edu \citep{fineweb-edu}, and NextCoder \citep{nextcoder}.
\begin{itemize}
    \item \textbf{Mixed text}: We compile a multilingual text corpus from diverse sources, including books, webpages, code, and published papers, to facilitate a comprehensive evaluation on LLMs' compression efficiency.
    \item \textbf{FinePDFs-en}: The FinePDFs dataset \citep{finepdfs} consists of about 3T tokens sourced exclusively from publicly available PDF files. We only select from the English subset to better examine the influence of the corpus distribution.
    \item \textbf{Ch-FineWeb-Edu}: The Chinese Fineweb Edu dataset \citep{fineweb-edu-ch} is a high-quality Chinese pretraining corpus of 90 million samples in the education domain, selected by a strategy similar to that of FineWeb-Edu.
    \item \textbf{FineWeb-Edu}: The FineWeb-Edu dataset \citep{fineweb-edu} contains 1.3T tokens of educational English webpages filtered from the FineWeb dataset, based on the annotations generated by Llama-3-70B-Instruct.
    \item \textbf{NextCoder}: The NextCoder dataset \citep{nextcoder} consists of 127K unique code samples generated by GPT-4o and Llama-3.3-70B-Instruct across 8 programming languages: Python, Java, C++, C, Rust, JavaScript, Go, and Kotlin.
\end{itemize}
We select samples that are sufficiently long from these datasets, such that the token length is larger than the truncation threshold $L$ for all evaluated models.
Table~\ref{tab:dataset} shows the details of the evaluation datasets regarding sample count, minimal sample length, average sample length, and offset value.

\begin{table*}[htb]
    \centering
    \resizebox{\columnwidth}{!}{
    \begin{NiceTabular}{lrccc}
        \toprule
        \textbf{Dataset} & \textbf{Sample count} & \textbf{Min sample length} & \textbf{Avg sample length} & \textbf{Offset value} \\
        \midrule
        Mixed text (Ours) & 200,000 & 1067 & 7735.8 & -24 \\
        FinePDFs-en \citep{finepdfs} & 161,102 & 1241 & 9349.2 & -27 \\
        Ch-FineWeb-Edu \citep{fineweb-edu-ch} & 116,090 & 1201 & 2150.7 & -18.7 \\
        FineWeb-Edu \citep{fineweb-edu} & 155,670 & 1201 & 3089.3 & -27 \\
        NextCoder \citep{nextcoder} & 79,965 & 1024 & 1391.8 & -27 \\
        \bottomrule
    \end{NiceTabular}
    }
    \caption{\textbf{Details of the evaluation datasets.} The sample length is evaluated using the DeepSeek-V3.1's highly efficient tokenizer.}
    \label{tab:dataset}
\end{table*}

\section{Results}
\subsection{Main Results} \label{sec:main_results}

As shown in Figure~\ref{fig:ic_mixed}, a series of models with varying sizes exhibit consistent information capacity, and thus we report the average information capacity for each model series.
Previous studies have established that the correlation between compression and intelligence weakens when the evaluation corpus significantly deviates from the domain of downstream tasks \citep{huang2024compressionrepresentsintelligencelinearly}.
Consequently, we select five heterogeneous evaluation datasets to comprehensively measure model efficiency, among which a mixed text corpus is employed to demonstrate overall intelligence.
Table~\ref{tab:ic_main_results} shows the evaluation results of the information capacity for mainstream open-source models across five datasets: Mixed text, FinePDFs-en \citep{finepdfs}, Ch-FineWeb-Edu \citep{fineweb-edu-ch}, FineWeb-Edu \citep{fineweb-edu}, and NextCoder \citep{nextcoder}.
The latest MoE models, exemplified by DeepSeek-V3.1 \citep{deepseekai2025deepseekv3technicalreport} and GLM-4.5 \citep{5team2025glm45agenticreasoningcoding}, achieve the highest information capacity on multiple datasets, followed by the latest dense models such as Qwen3 \citep{yang2025qwen3technicalreport}, Hunyuan, and GLM-4 \citep{glm2024chatglmfamilylargelanguage}.
These results align with their remarkable performance on downstream tasks other than text compression.

\begin{table*}[htb]
    \centering
    \resizebox{\columnwidth}{!}{
    \begin{NiceTabular}{c|cc|cc|cc|cc|cc}
       \toprule
       \multirow{2}{*}{\textbf{Model series}} & \multicolumn{2}{c}{\textbf{Mixed text}} & \multicolumn{2}{c}{\textbf{FinePDFs-en}} & \multicolumn{2}{c}{\textbf{Ch-FineWeb-Edu}} & \multicolumn{2}{c}{\textbf{FineWeb-Edu}} & \multicolumn{2}{c}{\textbf{NextCoder}} \\
       & \textbf{Rank} & \textbf{IC} $\uparrow$ & \textbf{Rank} & \textbf{IC} $\uparrow$ & \textbf{Rank} & \textbf{IC} $\uparrow$ & \textbf{Rank} & \textbf{IC} $\uparrow$ & \textbf{Rank} & \textbf{IC} $\uparrow$ \\
       \midrule
       GLM-4.5-Base       & \cellcolor{gray!30} 1 & \cellcolor{gray!30} 0.2415 & 2 & 0.2049 & 3 & 0.3411 & 2 & 0.2545 & 3 & 0.2163 \\
       DeepSeek-V3.1-Base & 2 & 0.2396 & \cellcolor{gray!30} 1 & \cellcolor{gray!30} 0.2196 & 2 & 0.3534 & \cellcolor{gray!30} 1 & \cellcolor{gray!30} 0.2571 & 8 & 0.1537 \\
       Hunyuan-Pretrain   & 3 & 0.2331 & 5 & 0.1916 & \cellcolor{gray!30} 1 & \cellcolor{gray!30} 0.3866 & 4 & 0.2348 & 7 & 0.1898 \\
       GLM-4-Base         & 4 & 0.2267 & 4 & 0.1939 & 4 & 0.3209 & 5 & 0.2326 & 4 & 0.2134 \\
       Qwen3-Base (Dense) & 5 & 0.2056 & 8 & 0.1664 & 8 & 0.2784 & 7 & 0.2165 & \cellcolor{gray!30} 1 & \cellcolor{gray!30} 0.2204 \\
       InternLM2.5        & 6 & 0.1900 & 9 & 0.1512 & 5 & 0.3023 & 11 & 0.1876 & 6 & 0.1955 \\
       Llama-4            & 7 & 0.1848 & 6 & 0.1880 & 10 & 0.1791 & 8 & 0.2071 & 5 & 0.1980 \\
       Seed-OSS-Base      & 8 & 0.1763 & 10 & 0.1488 & 6 & 0.2893 & 10 & 0.1978 & 9 & 0.0796\\
       Llama-3            & 9 & 0.1758 & 3 & 0.2041 & 11 & 0.0533 & 3 & 0.2388 & 2 & 0.2186 \\
       DeepSeek-V2        & 10 & 0.1672 & 11 & 0.1256 & 7 & 0.2847 & 9 & 0.2032 & 11 & 0.0260 \\
       Gemma-3-Base       & 11 & 0.1584 & 7 & 0.1764 & 9 & 0.1838 & 6 & 0.2222 & 10 & 0.0386 \\
       \bottomrule
    \end{NiceTabular}
    }
    \caption{\textbf{Information capacity leaderboard of mainstream open-source models.} The rankings of model series vary across datasets, indicating imbalanced capabilities in approximating probability distributions of heterogeneous text sources. Notably, the rankings significantly change across languages, which reveals strong linguistic biases in open-source LLMs. Note that numerical comparison of information capacities can only be conducted across models within a dataset (column-wise), since the offset value may vary across datasets, as shown in Table~\ref{tab:dataset}. (IC: Information capacity.)}
    \label{tab:ic_main_results}
\end{table*}

The rankings of model series vary across datasets, indicating imbalanced capabilities on heterogeneous text sources in terms of both probability estimation and downstream task performance.
Notably, the rankings differ significantly across languages, which reveals strong linguistic biases in open-source LLMs.
For instance, the Llama series from Meta \citep{grattafiori2024llama3herdmodels} and the Gemma series from Google \citep{gemmateam2025gemma3technicalreport} both perform poorly on the Chinese corpus from the Ch-FineWeb-Edu dataset, compared to other models developed by Chinese companies.
The rankings of information capacity also change when evaluated on computer code from the NextCoder dataset instead of English text from the FinePDFs-en dataset, even though the code is expressed in English characters.
Both FinePDFs-en and FineWeb-Edu datasets are in English, but the text sources are PDF files and webpages, respectively, which also cause slight variations in model rankings.
These findings highlight the necessity of a holistic approach in model training across different languages and text sources to ensure consistent performance.

\subsection{Empirical Findings}
\subsubsection{Tokenizer Efficiency}

\begin{figure}[htb]
    \centering
    \subfloat[Mixed text]{\includegraphics[width=.5\linewidth]{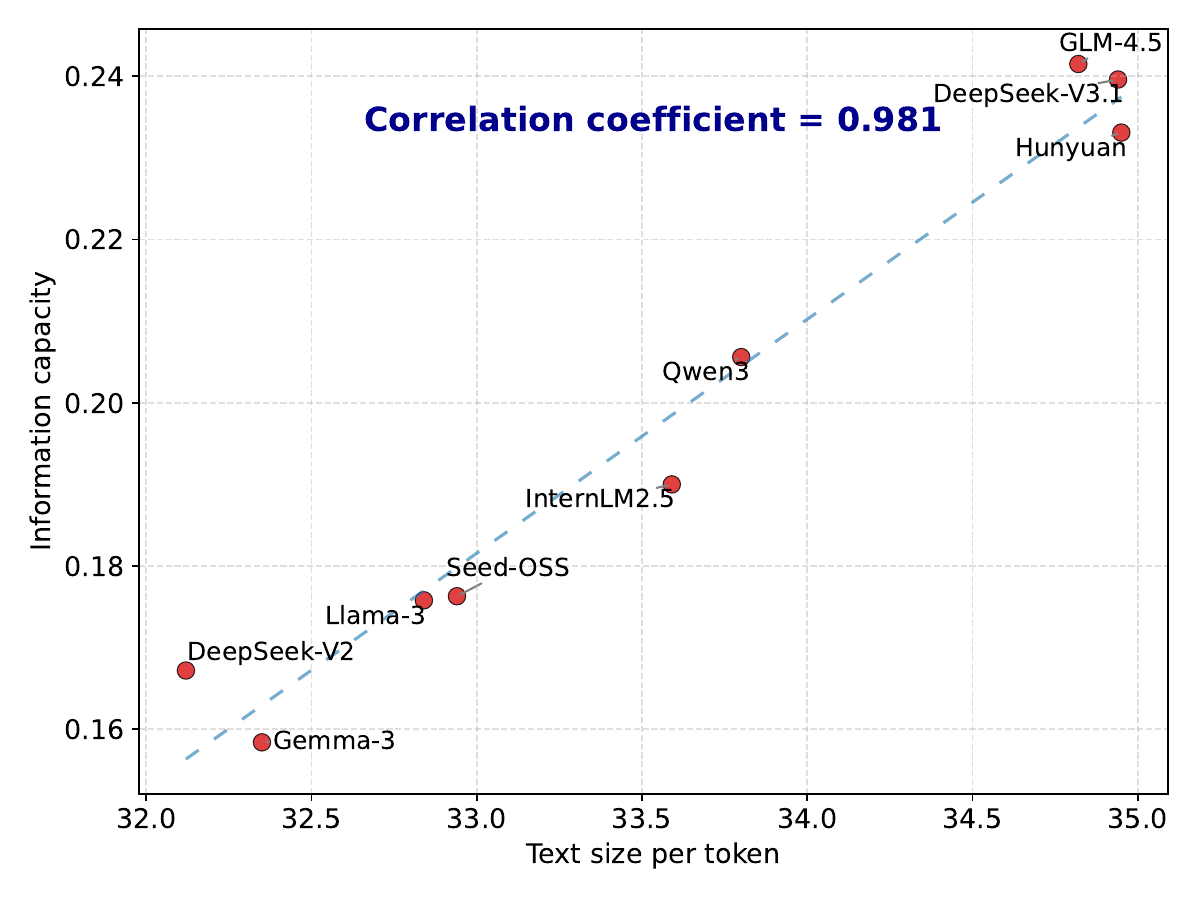}}
    \subfloat[FinePDFs-en]{\includegraphics[width=.5\linewidth]{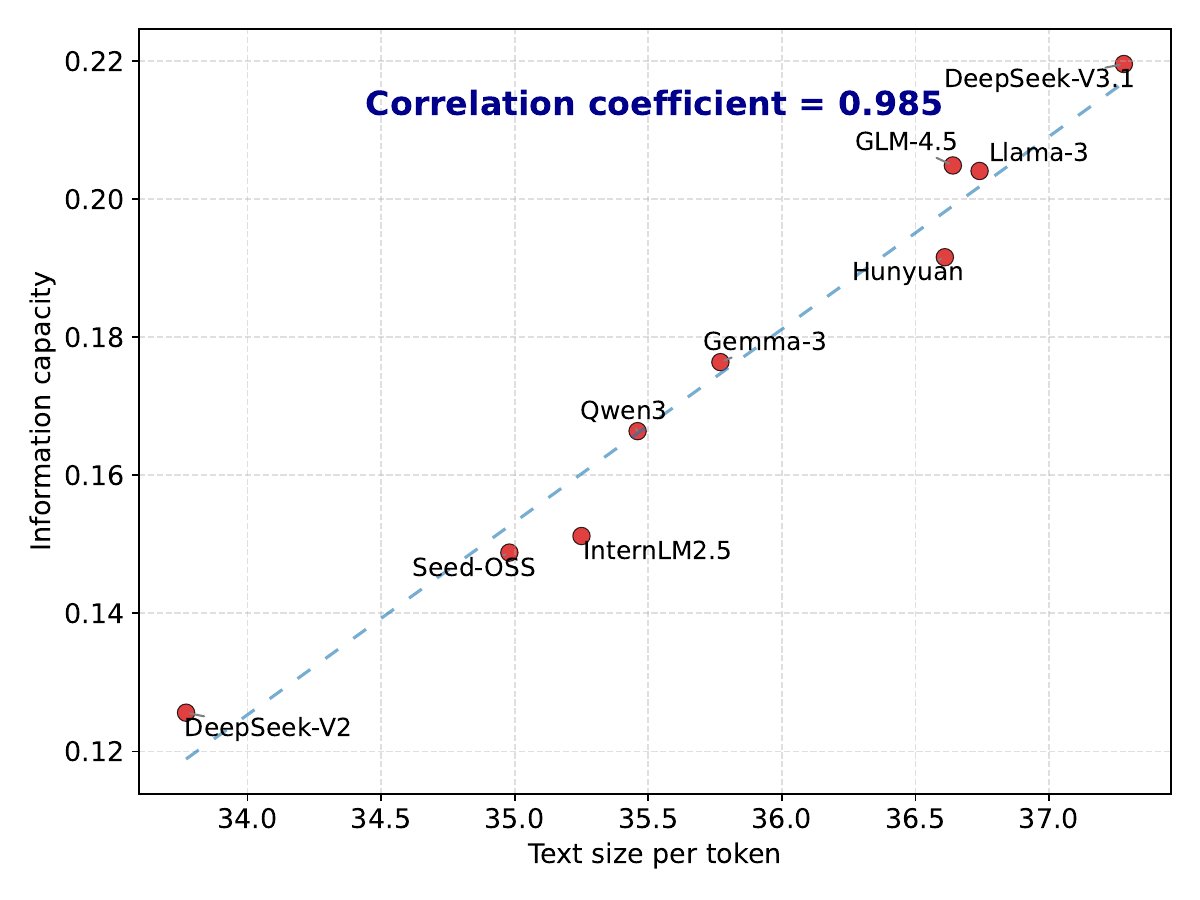}} \hfill
    \subfloat[Ch-FineWeb-Edu]{\includegraphics[width=.5\linewidth]{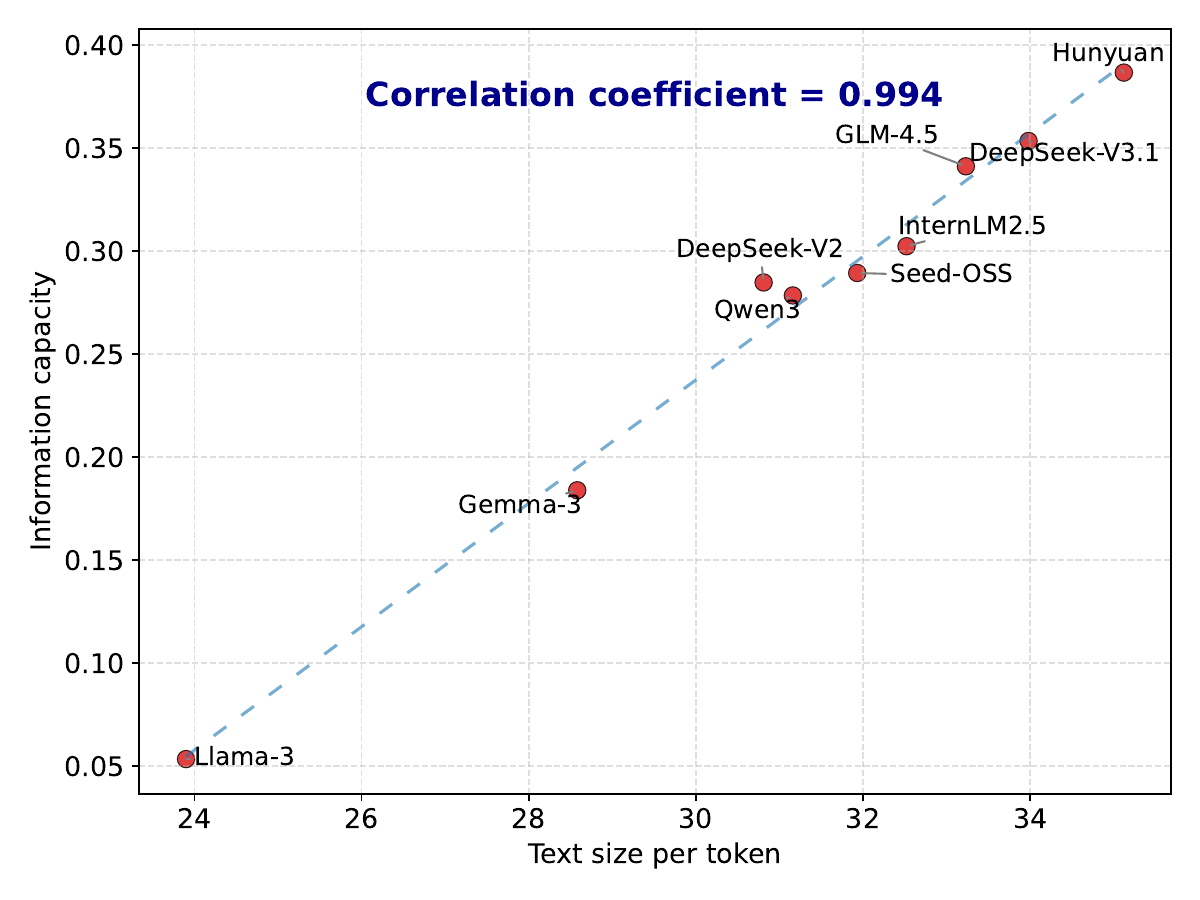}}
    \subfloat[NextCoder]{\includegraphics[width=.5\linewidth]{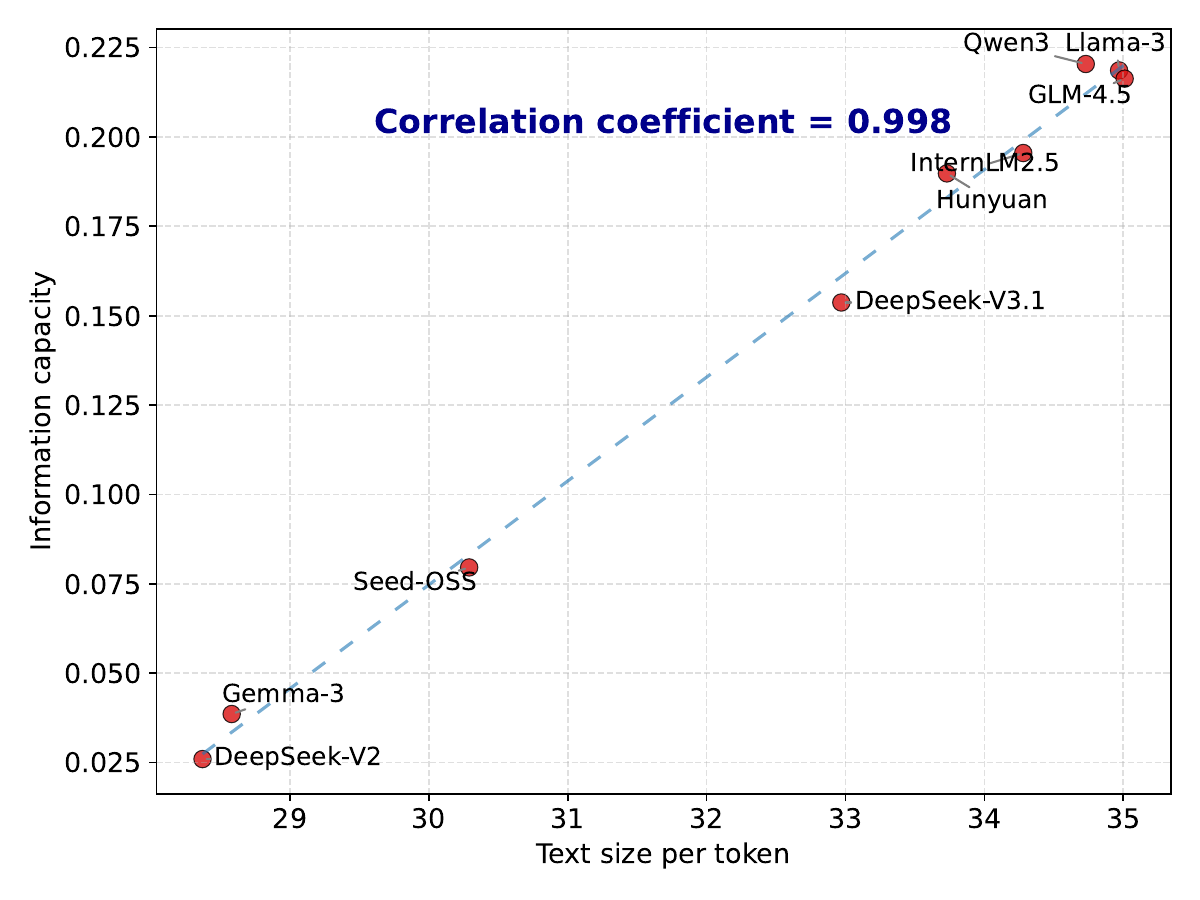}}
    \caption{\textbf{Impact of tokenizer efficiency on information capacity.} The information capacity scales almost linearly with the average text size per token across multiple datasets, with Pearson correlation coefficients consistently exceeding 0.98.}
    \label{fig:tokenizer}
\end{figure}

We find that tokenizer efficiency is the dominant factor in information capacity.
Figure~\ref{fig:tokenizer} demonstrates the impact of tokenizer efficiency on information capacity across four datasets: Mixed text, FinePDFs-en \citep{finepdfs}, Ch-FineWeb-Edu \citep{fineweb-edu-ch}, and NextCoder \citep{nextcoder}.
Results show that the information capacity scales almost linearly with the average text size per token across multiple datasets, with Pearson correlation coefficients consistently exceeding 0.98.
When testing on different datasets with distinct distributions, the average text size per token for a specific LLM varies significantly, and the rankings of tokenizer efficiencies across different LLMs change correspondingly.
This phenomenon is also observed in previous studies on tokenizer performance \citep{tamang2024evaluatingtokenizerperformancelarge}.
However, the strong linear correlation between the information capacity and the average text size per token persists across all evaluated datasets.

Quantitative comparison highlights the significance of tokenizer efficiency in an LLM's compression capability.
In our mixed dataset, the average text size per token for the latest LLMs using the universal UTF-8 encoding varies from 32.35 bits (Gemma-3) to 34.94 bits (DeepSeek-V3.1), with a range of 2.59 bits.
Conversely, the average NLL per token for different models with similar sizes, which corresponds to the symbol length after arithmetic coding, exhibits much smaller variations.
For models with 7 billion to 8 billion parameters, the NLL ranges from 2.822 bits (Llama-3.1-8B) to 3.155 bits (InternLM2.5-7B), with a difference of 0.333 bits.
As a result, the tokenizer should be carefully designed in order to maximize the LLM's compression efficiency.

\subsubsection{Pretraining Data}

\begin{table*}[htb]
    \centering
    \begin{NiceTabular}{cc|cc|cc|cc}
       \toprule
       \multirow{2}{*}{\textbf{Model}} & \multirow{2}{*}{\textbf{PT (T)}} & \multicolumn{2}{c}{\textbf{Mixed text}} & \multicolumn{2}{c}{\textbf{FinePDFs-en}} & \multicolumn{2}{c}{\textbf{Ch-FineWeb-Edu}} \\
       & & \textbf{NLL} $\downarrow$ & \textbf{IC} $\uparrow$ & \textbf{NLL} $\downarrow$ & \textbf{IC} $\uparrow$ & \textbf{NLL} $\downarrow$ & \textbf{IC} $\uparrow$ \\
       \midrule
       \multirow{6}{*}{\centering TinyLlama-1.1B} &
       0.5 & 3.290 & \multicolumn{1}{l}{0.0681} & 3.198 & \multicolumn{1}{l}{0.0904} & 4.481 & \multicolumn{1}{l}{0.0168} \\
       & 1 & 2.966 & 0.0785 {\textcolor{upColor}{$\uparrow$0.0104}} & 3.036 & 0.0956 {\textcolor{upColor}{$\uparrow$0.0052}} & 3.260 & 0.0561 {\textcolor{upColor}{$\uparrow$0.0393}} \\
       & 1.5 & 2.935 & 0.0795 {\textcolor{upColor}{$\uparrow$0.0010}} & 3.009 & 0.0965 {\textcolor{upColor}{$\uparrow$0.0009}} & 3.197 & 0.0581 {\textcolor{upColor}{$\uparrow$0.0020}} \\
       & 2 & 2.916 & 0.0801 {\textcolor{upColor}{$\uparrow$0.0006}} & 2.987 & 0.0972 {\textcolor{upColor}{$\uparrow$0.0007}} & 3.166 & 0.0591 {\textcolor{upColor}{$\uparrow$0.0010}} \\
       & 2.5 & 2.758 & 0.0852 {\textcolor{upColor}{$\uparrow$0.0051}} & 2.849 & 0.1016 {\textcolor{upColor}{$\uparrow$0.0044}} & 2.919 & 0.0671 {\textcolor{upColor}{$\uparrow$0.0080}} \\
       & 3 & 2.739 & 0.0858 {\textcolor{upColor}{$\uparrow$0.0006}} & 2.836 & 0.1020 {\textcolor{upColor}{$\uparrow$0.0004}} & 2.885 & 0.0682 {\textcolor{upColor}{$\uparrow$0.0011}} \\
       \midrule
       Qwen2-0.5B & 7 & 3.674 & \multicolumn{1}{l}{0.2046} & 3.686 & \multicolumn{1}{l}{0.1594} & 4.184 & \multicolumn{1}{l}{0.2764} \\
       Qwen2.5-0.5B & 18 & 3.648 & 0.2055 {\textcolor{upColor}{$\uparrow$0.0009}} & 3.575 & 0.1631 {\textcolor{upColor}{$\uparrow$0.0037}} & 4.130 & 0.2782 {\textcolor{upColor}{$\uparrow$0.0018}} \\
       \midrule
       Qwen2-1.5B & 7 & 3.272 & \multicolumn{1}{l}{0.2068} & 3.273 & \multicolumn{1}{l}{0.1643} & 3.729 & \multicolumn{1}{l}{0.2766} \\
       Qwen2.5-1.5B & 18 & 3.238 & 0.2079 {\textcolor{upColor}{$\uparrow$0.0011}} & 3.165 & 0.1677 {\textcolor{upColor}{$\uparrow$0.0034}} & 3.651 & 0.2791 {\textcolor{upColor}{$\uparrow$0.0025}} \\
       \midrule
       Qwen2-72B & 7 & 2.518 & \multicolumn{1}{l}{0.1964} & 2.586 & \multicolumn{1}{l}{0.1585} & 2.811 & \multicolumn{1}{l}{0.2603} \\
       Qwen2.5-72B & 18 & 2.507 & 0.1967 {\textcolor{upColor}{$\uparrow$0.0003}} & 2.540 & 0.1597 {\textcolor{upColor}{$\uparrow$0.0012}} & 2.778 & 0.2612 {\textcolor{upColor}{$\uparrow$0.0009}} \\
       \bottomrule
    \end{NiceTabular}
    \caption{\textbf{Impact of pretraining data on information capacity.} With the increase in the pretraining dataset size, the NLL of the next token predicted by LLMs consistently decreases, and the information capacity improves correspondingly. The \textcolor{upColor}{green numbers} denote the gain in information capacity relative to the preceding model. (PT: Pretrained tokens, IC: Information capacity.)}
    \label{tab:pretrain_data}
\end{table*}

Another important factor in information capacity is the pretraining data.
Table~\ref{tab:pretrain_data} demonstrates the impact of pretraining dataset size on information capacity across three datasets: Mixed text, FinePDFs-en \citep{finepdfs}, and Ch-FineWeb-Edu \citep{fineweb-edu-ch}.
With the increase in the pretraining dataset size, the NLL of the next token predicted by LLMs consistently decreases, and the information capacity grows correspondingly.
For the TinyLlama-1.1B model \citep{zhang2024tinyllamaopensourcesmalllanguage}, the gain in information capacity resulting from additional pretraining with 0.5T tokens exhibits no consistent trend.
The additional pretraining from 0.5T tokens to 1T tokens and from 2T tokens to 2.5T tokens leads to a significant NLL reduction and thus a remarkable increase in information capacity, while others only bring slight gains.
These results are consistent across datasets and are presumably caused by the quality variation among different portions of the pretraining data.
Additionally, Qwen2 \citep{yang2024qwen2technicalreport} and Qwen2.5 \citep{qwen2025qwen25technicalreport} share identical model architecture, and the key difference for the base model under evaluation lies in the pretraining data.
Qwen2.5's pretraining dataset consists of about 18T tokens, with a significant increase from Qwen2's 7T tokens.
This increase in pretraining data only provides a slight information capacity gain, showing that additional training brings diminishing returns when the model has already been trained on sufficient high-quality data, which is also observed in previous studies on sub-optimal scaling \citep{scalingdataconstrained,chen-etal-2025-revisiting}.

\subsubsection{MoE Architecture}

Model architecture constitutes the last prominent factor in information capacity.
For example, the MLA \citep{deepseekai2024deepseekv2strongeconomicalefficient} used by DeepSeek's models can reduce the computational complexity of the attention mechanism for long token sequences, compared to the mainstream GQA \citep{ainslie2023gqatraininggeneralizedmultiquery}.
However, the evaluation of the model's information capacity employs a relatively small sequence length, where the attention mechanism only constitutes a small portion of inference FLOPs.
In addition, the computational complexity used in the calculation of information capacity is measured on a logarithmic scale.
Thus, the difference in attention implementations exerts a marginal influence on the model's information capacity.

In contrast, the MoE architecture \citep{shazeer2017outrageouslylargeneuralnetworks} drastically reduces inference FLOPs by only activating a small portion of total parameters.
As shown in Table~\ref{tab:moe}, the MoE architecture enhances the LLM's ability to predict the next token while maintaining low computational complexity.
Results on Qwen1.5 \citep{bai2023qwentechnicalreport} and Qwen2 \citep{yang2024qwen2technicalreport} show that the NLL of MoE models is comparable to that of dense variants with similar numbers of total parameters, while the FLOPs count is mainly determined by the number of activated parameters.
Consequently, the compression gain of MoE models is larger while maintaining similar computational complexity, compared to dense variants with similar activated parameters, thus achieving higher information capacity.
For MoE models, the sparsity ratio is defined as the number of activated parameters divided by the total parameter count.
A lower sparsity ratio further extends total parameters while maintaining an identical activated parameter count, hence a more significant gain in information capacity.
As exemplified by the Llama-4 models \citep{meta2025llama4}, both the 109B and 400B variants only activate 17B parameters, and the FLOPs counts are equal.
However, the 400B variant can better predict the next token with an average NLL of 3.907, lower than the 3.977 achieved by the 107B variant. 
Consequently, the information capacity of the 400B variant increases from 0.1836 to 0.1856 only due to a lower sparsity ratio, which is also the reason that the gain in information capacity achieved by the MoE architecture for the Qwen1.5 series is larger than that for the Qwen2 series.
These results are aligned with previous studies on the benefits of a higher sparsity ratio to the scaling performance \citep{wang-etal-2024-scaling,abnar2025parametersvsflopsscaling,tian2025greaterleveragescalinglaws}.

\begin{table*}[htb]
    \centering
    \begin{NiceTabular}{cccccc}
       \toprule
       \textbf{Model} & \textbf{Total params (B)} & \textbf{Activated params (B)} & \textbf{NLL} $\downarrow$ & \textbf{FLOPs (G)} & \textbf{IC} $\uparrow$ \\
       \midrule
       \multirow{4}{*}{\centering Qwen1.5} &
       1.8 & 1.8 (100\%) & 3.419 & 3.151 & 0.2022 \\
       & 4 & 4 (100\%) & 3.175 & 7.331 & 0.2022 \\
       & 14 & 14 (100\%) & 2.944 & 27.194 & 0.1978 \\
       & 32 & 32 (100\%) & 2.716 & 64.135 & 0.1973 \\
       \rowcolor{gray!20} \textbf{Qwen1.5-MoE} & \textbf{14.3} & \textbf{2.7 (18.9\%)} & \textbf{2.895} & \textbf{4.850} & \textbf{0.2146} \\
       \midrule
       \multirow{4}{*}{\centering Qwen2}
       & 0.5 & 0.5 (100\%) & 3.674 & 1.032 & 0.2046 \\
       & 1.5 & 1.5 (100\%) & 3.272 & 3.175 & 0.2068 \\
       & 7 & 7 (100\%) & 2.886 & 14.346 & 0.2049 \\
       & 72 & 72 (100\%) & 2.518 & 144.258 & 0.1964 \\
       \rowcolor{gray!20} \textbf{Qwen2-MoE} & \textbf{57} & \textbf{14 (24.6\%)} & \textbf{2.668} & \textbf{26.676} & \textbf{0.2059} \\
       \midrule
       \rowcolor{gray!20} \multirow{2}{*}{\centering \textbf{Llama-4 (MoE)}}
        & 109 & 17 (15.6\%) & 3.977 & 40.824 & 0.1836 \\
       \rowcolor{gray!20} & \textbf{400} & \textbf{17 (4.25\%)} & \textbf{3.897} & \textbf{40.824} & \textbf{0.1859} \\
       \bottomrule
    \end{NiceTabular}
    \caption{\textbf{Impact of MoE architecture on information capacity.} The MoE architecture enhances the LLM's ability to predict the next token while maintaining low computational complexity. A lower sparsity ratio further extends total parameters, leading to a more significant gain in information capacity. (IC: Information capacity.)}
    \label{tab:moe}
\end{table*}

\subsection{Ablation Study}
\subsubsection{Impact of Post-training}

Modern LLMs are trained in two separate stages, namely pre-training and post-training.
The first stage trains the model to predict the next token on the text corpus, which directly corresponds to the text compression task.
Conversely, the second stage aims to improve the model's capabilities in instruction following and advanced reasoning, so that it becomes a helpful assistant capable of responding to user requests in a conversational format.
This post-training enhances user experience and model performance on other downstream tasks, but at the cost of degraded capability in modeling the conditional probability on the text corpus.
However, the pretraining loss still provides a reliable precursor for the downstream performance of the instruction-tuned model that has gone through the second stage training \citep{loss_and_task_performance1,loss_and_task_performance2}.

\begin{figure}[htb]
    \centering
    \vspace{-6pt}
    \includegraphics[width=.75\linewidth]{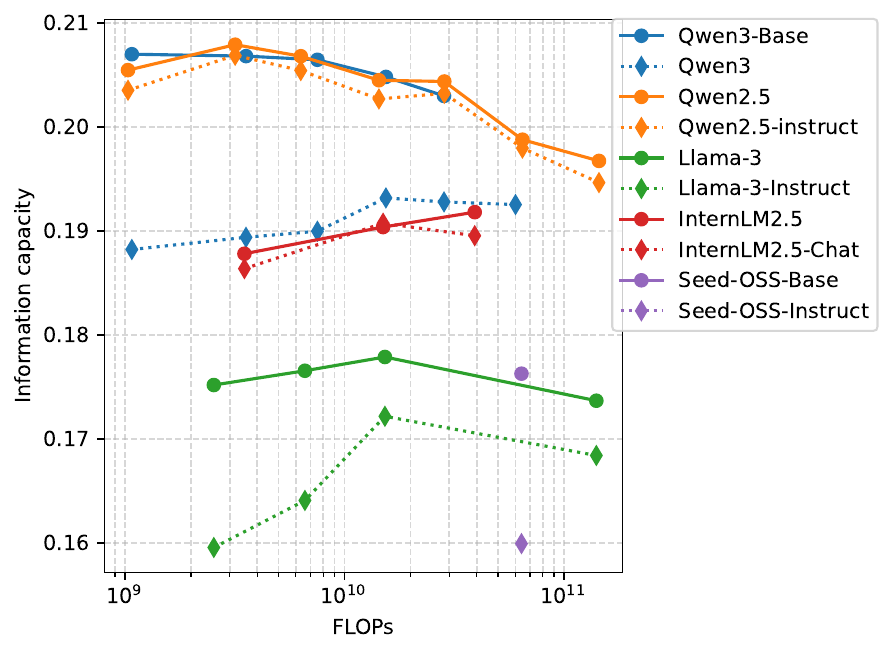}
    \vspace{-6pt}
    \caption{\textbf{Impact of post-training on information capacity.} Post-training impairs LLMs' capability in predicting the next token for plain text, thus degrading the information capacity. Latest LLMs utilize sophisticated post-training methods, which cause more severe degradations in compression performance.}
    \label{fig:post_training}
\end{figure}

Figure~\ref{fig:post_training} compares the information capacity of different models before and after post-training, measured on our mixed text dataset.
Results show that the post-training of modern LLMs impairs the model's capability in predicting the next token for plain text, degrading the text compression efficiency and the information capacity.
The latest LLMs utilize sophisticated post-training methods in addition to supervised fine-tuning (SFT).
For instance, the Qwen3 series \citep{yang2025qwen3technicalreport} employs multi-stage reinforcement learning (RL) to grant LLMs advanced capabilities such as hybrid thinking and tool calling.
However, these techniques cause more severe degradations in compression performance.

To ensure a fair comparison and avoid the interference caused by post-training, we evaluate the checkpoints before the second stage training.
As a number of influential base models are not released to the public, exemplified by gpt-oss from OpenAI \citep{openai2025gptoss120bgptoss20bmodel}, Phi-4 from Microsoft \citep{abdin2024phi4technicalreport}, and the MoE variants of the Qwen3 series, it is unable to evaluate their information capacity accurately.
Additionally, the pretraining corpora of some open-source base models incorporate synthetic instruction tuning data, apart from large-scale plain text normally used during the first-stage training.
In Seed-OSS model series, a variant of the base model trained without instruction data is also released, denoted as Seed-OSS-36B-Base-woSyn.
We observe a marginal decrease in the measured information capacity from 0.17633 to 0.17628 after incorporating these instruction data, albeit performance improvements on some benchmarks \citep{seed2025}.
In contrast, the Instruct variant has an information capacity of 0.15995, significantly lower than that of the Base variant, which further verifies our claim that more advanced RL methods cause more severe degradations.
As other model series do not release the weights of base models trained without instruction data, despite the potential usage of such data, we report results of the standard base models to maintain consistency.

\subsubsection{Impact of Test Sample Length}

\begin{table*}[htb]
    \centering
    \begin{NiceTabular}{c|cccc|cccc}
       \toprule
       \multirow{2}{*}{\textbf{Model}} & \multicolumn{4}{c}{$L = 1024$} & \multicolumn{4}{c}{$L = 2048$} \\
       & \textbf{TS} $\uparrow$ & \textbf{NLL} $\downarrow$ & \textbf{FLOPs (G)} & \textbf{IC} $\uparrow$ & \textbf{TS} $\uparrow$ & \textbf{NLL} $\downarrow$ & \textbf{FLOPs (G)} & \textbf{IC} $\uparrow$ \\
       \midrule
       Qwen3-0.6B-Base & 33.80 & 3.590 & 1.074 & 0.2070 & 33.78 & 3.526 & 1.133 & 0.2079 {\textcolor{upColor}{$\uparrow$0.0009}} \\
       Qwen3-1.7B-Base & 33.80 & 3.238 & 3.558 & 0.2068 & 33.78 & 3.180 & 3.676 & 0.2077 {\textcolor{upColor}{$\uparrow$0.0009}} \\
       Qwen3-4B-Base   & 33.80 & 3.026 & 7.525 & 0.2065 & 33.78 & 2.969 & 7.714 & 0.2074 {\textcolor{upColor}{$\uparrow$0.0009}} \\
       Qwen3-8B-Base   & 33.80 & 2.868 & 15.438 & 0.2048 & 33.78 & 2.815 & 15.740 & 0.2056 {\textcolor{upColor}{$\uparrow$0.0008}} \\
       Qwen3-14B-Base  & 33.80 & 2.751 & 28.399 & 0.2030 & 33.78 & 2.700 & 28.818 & 0.2038 {\textcolor{upColor}{$\uparrow$0.0008}} \\
       \midrule
       Llama-3.2-1B & 32.84 & 3.367 & 2.539 & 0.1752 & 32.80 & 3.287 & 2.606 & 0.1763 {\textcolor{upColor}{$\uparrow$0.0011}} \\
       Llama-3.2-3B & 32.84 & 3.081 & 6.601 & 0.1766 & 32.80 & 3.002 & 6.777 & 0.1775 {\textcolor{upColor}{$\uparrow$0.0009}} \\
       Llama-3.1-8B & 32.84 & 2.822 & 15.277 & 0.1779 & 32.80 & 2.748 & 15.546 & 0.1788 {\textcolor{upColor}{$\uparrow$0.0009}} \\
       \midrule
       GLM-4-9B-Base  & 34.82 & 3.027 & 17.893 & 0.2288 & 34.77 & 2.927 & 18.228 & 0.2301 {\textcolor{upColor}{$\uparrow$0.0013}} \\
       GLM-4-32B-Base & 34.82 & 2.761 & 64.034 & 0.2245 & 34.77 & 2.697 & 64.801 & 0.2248 {\textcolor{upColor}{$\uparrow$0.0003}} \\
       \midrule
       Seed-OSS-36B-Base & 32.94 & 2.612 & 63.999 & 0.1763 & 32.89 & 2.549 & 64.670 & 0.1766 {\textcolor{upColor}{$\uparrow$0.0003}} \\
       \bottomrule
    \end{NiceTabular}
    \caption{\textbf{Impact of test sample length on information capacity.} When the text length increases, the LLM predicts these newly added tokens using an extended context from the preceding text. Consequently, the NLL of the next token predicted by LLMs slightly decreases while the average required FLOPs per token marginally grows. The information capacity marginally improves when the text length increases from 1024 to 2048, as denoted by the \textcolor{upColor}{green numbers} in the IC column of $L = 2048$. (TS: Text size, IC: Information capacity.)}
    \label{tab:sample_length}
\end{table*}

Table~\ref{tab:sample_length} shows the impact of test sample length on information capacity, evaluated on our mixed text dataset.
When the text length grows, the LLM predicts these newly added tokens using an extended context from the preceding text.
As a result, the LLM provides a more accurate prediction on the following content, leading to a lower average NLL and a higher compression gain.
Additionally, the extended context introduces more computational overhead to the attention mechanisms inherent in LLMs, and the average required FLOPs per token marginally increases.
These two factors cause a marginal increase in the information capacity on Qwen3 \citep{yang2025qwen3technicalreport}, Llama-3 \citep{grattafiori2024llama3herdmodels}, GLM-4 \citep{glm2024chatglmfamilylargelanguage}, and Seed-OSS models \citep{seed2025} when the text length increases from 1024 to 2048.
However, the overall difference caused by test sample length is negligible compared to that between different models.

\subsubsection{Impact of Softmax Temperature}

\begin{figure}[!htb]
    \centering
    \includegraphics[width=.8\linewidth]{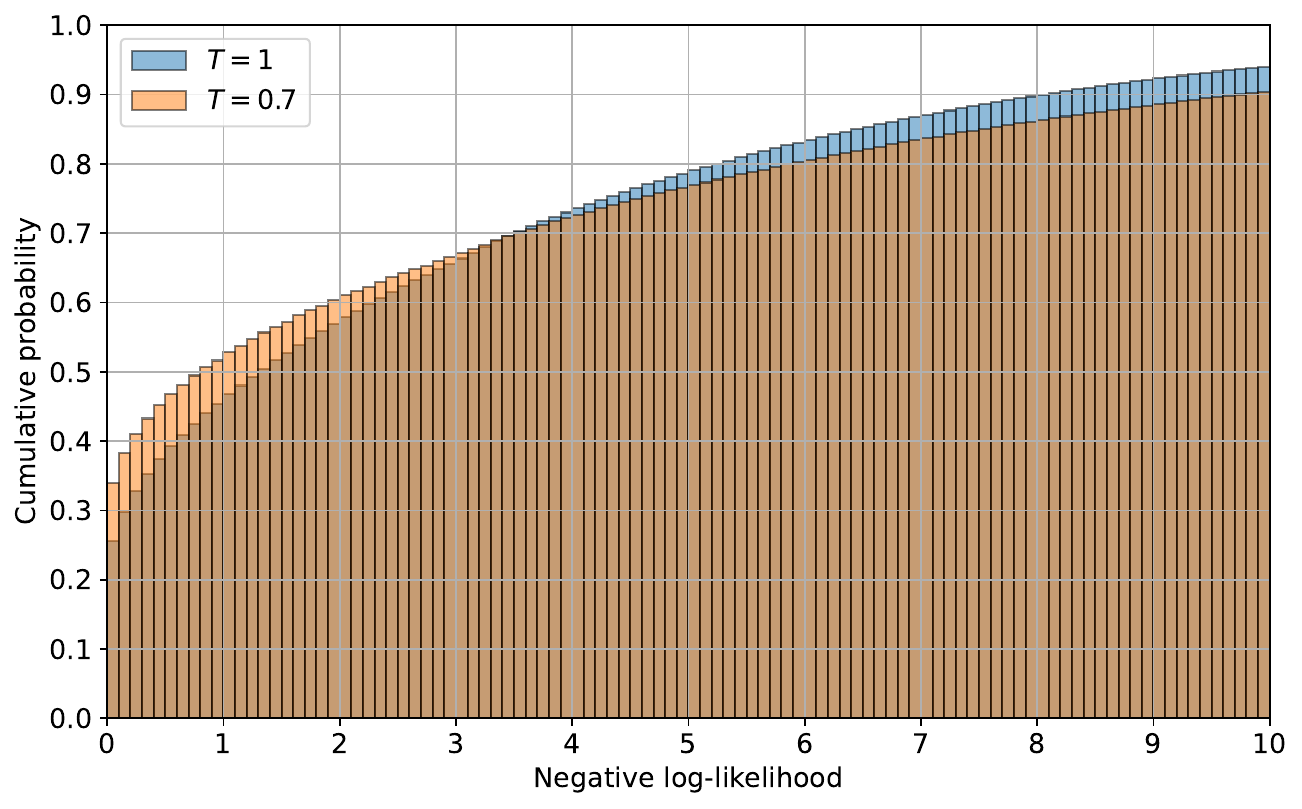}
    \caption{\textbf{Impact of softmax temperature on the cumulative probability of NLL.} A low temperature concentrates estimated probabilities on high-valued logits, which reduces the NLL when the top prediction on the next token is correct, but aggravates NLL penalties for errors. Consequently, a balanced temperature value minimizes the overall NLL, thereby maximizing the information capacity.}
    \label{fig:pdf_temperature}
\end{figure}

The temperature value used in the softmax function controls the shape of the probability distribution estimated by an LLM.
Figure~\ref{fig:pdf_temperature} demonstrates the impact of softmax temperature on the cumulative probability distribution of NLL, measured with the Qwen3-8B-Base model on our mixed text dataset.
A low temperature concentrates estimated probabilities on high-valued logits, which reduces the NLL when the top prediction on the next token is correct, but aggravates NLL penalties for errors. 
Consequently, a balanced temperature value minimizes the overall NLL, thereby maximizing the information capacity.

\begin{table*}[!htb]
    \centering
    \begin{NiceTabular}{c|cc|cc|cc}
       \toprule
       \multirow{2}{*}{\textbf{Model}} & \multicolumn{2}{c}{$T = 0.7$} & \multicolumn{2}{c}{$T = 1$} & \multicolumn{2}{c}{$T = 1.3$} \\
       & \textbf{NLL} $\downarrow$ & \textbf{IC} $\uparrow$ & \textbf{NLL} $\downarrow$ & \textbf{IC} $\uparrow$ & \textbf{NLL} $\downarrow$ & \textbf{IC} $\uparrow$ \\
       \midrule
       Qwen3-0.6B-Base & 3.972 & 0.1943 {\textcolor{downColor}{$\downarrow$0.0127}} & 3.590 & 0.2070 & 3.820 & 0.1993 {\textcolor{downColor}{$\downarrow$0.0077}} \\
       Qwen3-1.7B-Base & 3.579 & 0.1961 {\textcolor{downColor}{$\downarrow$0.0107}} & 3.238 & 0.2068 & 3.447 & 0.2002 {\textcolor{downColor}{$\downarrow$0.0066}} \\
       Qwen3-4B-Base & 3.345 & 0.1967 {\textcolor{downColor}{$\downarrow$0.0098}} & 3.026 & 0.2065 & 3.222 & 0.2005 {\textcolor{downColor}{$\downarrow$0.0060}} \\
       Qwen3-8B-Base & 3.169 & 0.1959 {\textcolor{downColor}{$\downarrow$0.0089}} & 2.868 & 0.2048 & 3.048 & 0.1995 {\textcolor{downColor}{$\downarrow$0.0053}} \\
       Qwen3-14B-Base & 3.040 & 0.1947 {\textcolor{downColor}{$\downarrow$0.0083}} & 2.751 & 0.2030 & 2.921 & 0.1981 {\textcolor{downColor}{$\downarrow$0.0049}} \\
       \midrule
       Llama-3.2-1B & 3.720 & 0.1639 {\textcolor{downColor}{$\downarrow$0.0113}} & 3.367 & 0.1752 & 3.604 & 0.1676 {\textcolor{downColor}{$\downarrow$0.0076}} \\
       Llama-3.2-3B & 3.411 & 0.1664 {\textcolor{downColor}{$\downarrow$0.0102}} & 3.081 & 0.1766 & 3.299 & 0.1699 {\textcolor{downColor}{$\downarrow$0.0067}} \\
       Llama-3.1-8B & 3.116 & 0.1692 {\textcolor{downColor}{$\downarrow$0.0087}} & 2.822 & 0.1779 & 3.015 & 0.1722 {\textcolor{downColor}{$\downarrow$0.0057}} \\
       \midrule
       GLM-4-9B-Base & 3.370 & 0.2188 {\textcolor{downColor}{$\downarrow$0.0100}} & 3.027 & 0.2288 & 3.231 & 0.2228 {\textcolor{downColor}{$\downarrow$0.0060}} \\
       GLM-4-32B-Base & 3.047 & 0.2165 {\textcolor{downColor}{$\downarrow$0.0080}} & 2.761 & 0.2245 & 2.961 & 0.2189 {\textcolor{downColor}{$\downarrow$0.0056}} \\
       \midrule
       Hunyuan-0.5B-Pretrain & 4.433 & 0.2182 {\textcolor{downColor}{$\downarrow$0.0149}} & 3.989 & 0.2331 & 4.231 & 0.2250 {\textcolor{downColor}{$\downarrow$0.0081}} \\
       Hunyuan-1.8B-Pretrain & 3.998 & 0.2187 {\textcolor{downColor}{$\downarrow$0.0130}} & 3.585 & 0.2317 & 3.793 & 0.2251 {\textcolor{downColor}{$\downarrow$0.0066}} \\
       Hunyuan-4B-Pretrain & 3.667 & 0.2213 {\textcolor{downColor}{$\downarrow$0.0112}} & 3.298 & 0.2325 & 3.499 & 0.2264 {\textcolor{downColor}{$\downarrow$0.0059}} \\
       Hunyuan-7B-Pretrain & 3.336 & 0.2251 {\textcolor{downColor}{$\downarrow$0.0100}} & 2.998 & 0.2351 & 3.178 & 0.2297 {\textcolor{downColor}{$\downarrow$0.0054}} \\
       \bottomrule
    \end{NiceTabular}
    \caption{\textbf{Impact of softmax temperature on information capacity.} The information capacity decreases when the temperature value deviates from its default value $T = 1$. The \textcolor{downColor}{red numbers} denote the reduction in information capacity relative to the case when $T = 1$. (IC: Information capacity.)}
    \label{tab:ic_temperature}
\end{table*}

Table~\ref{tab:ic_temperature} shows the impact of softmax temperature on information capacity.
When the temperature value deviates from its default value $T = 1$ used in all previous evaluations, the average NLL increases and the information capacity reduces correspondingly.
This trend persists across multiple model series, including Qwen3 \citep{yang2025qwen3technicalreport}, Llama-3 \citep{grattafiori2024llama3herdmodels}, GLM-4 \citep{glm2024chatglmfamilylargelanguage}, and Hunyuan.
Among these model series, only Hunyuan designates a temperature value of $T = 0.7$ in its \texttt{generation\_config.json} file, and others use the default temperature $T = 1$.
However, the default temperature $T = 1$ still significantly outperforms $T = 0.7$ when evaluating the information capacity of Hunyuan models, which validates our claim in Section~\ref{sec:evaluation_method}.

\subsection{Performance Prediction}

One of the applications of information capacity is performance prediction across different-sized models that belong to a series.
Assuming that the information capacity calculated from (\ref{eq:IC_biased}) is identical for a model series, the NLL performance of a target model can be estimated based on its size and true test results on a reference model, given by:
\begin{equation}
    \text{NLL}_{\text{target}} = C + b - (C + b - \text{NLL}_{\text{ref}}) \frac{\log_2 N_{\text{target}}}{\log_2 N_{\text{ref}}} ,
    \label{eq:NLL_prediction}
\end{equation}
where $\text{NLL}_{\text{target}}$ and $\text{NLL}_{\text{ref}}$ denote average NLLs on the given data samples for the target and reference models, respectively, and $N_{\text{target}}$ and $N_{\text{ref}}$ denote average inference FLOPs per token for the target and reference models, respectively.
This NLL estimation method can be easily extended to the pretraining scenario, simply by replacing the average NLL measured on given text with the cross-entropy loss at the final stage of pretraining.
The rationality of this extension stems from the uniformity between NLL and cross-entropy loss, as detailed in Section~\ref{sec:llm_probability_estimator}.
Notably, only one reference model is required to be trained and evaluated for performance estimation of pretrained models with arbitrary sizes, based on the consistent information capacity among a model series.
Consequently, the pretraining loss of an enormous model can be predicted from the loss of a substantially smaller reference model, thus accelerating the development of LLM pretraining.
In contrast, recent works on the scaling law of LLMs \citep{chinchilla_scaling_law, sardana2025chinchillaoptimalaccountinginferencelanguage,chen-etal-2025-revisiting,tian2025greaterleveragescalinglaws} typically estimate multiple coefficients from a large number of data points, consuming enormous computing resources.

\begin{figure}[htb]
    \centering
    \subfloat[Qwen3 series]{\includegraphics[width=.5\linewidth]{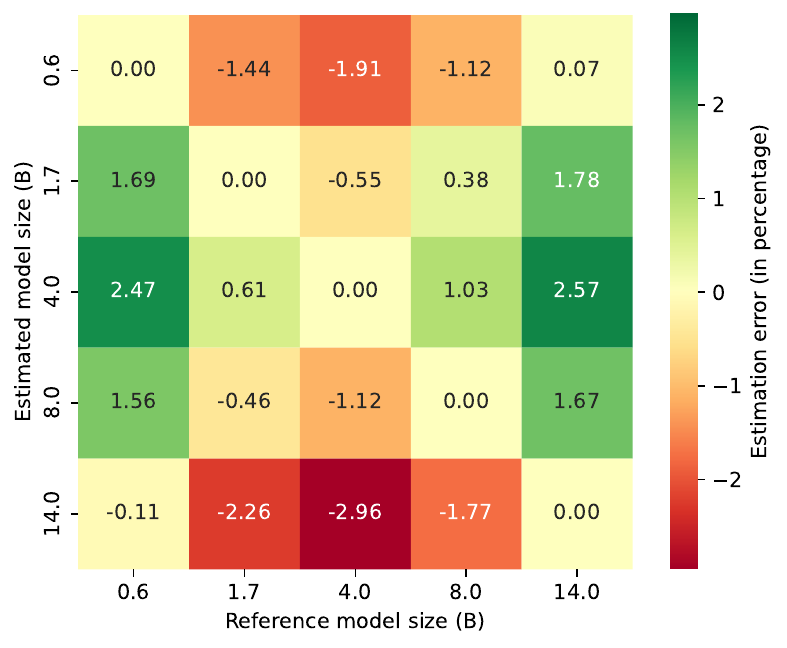}}
    \subfloat[Qwen1.5 series]{\includegraphics[width=.5\linewidth]{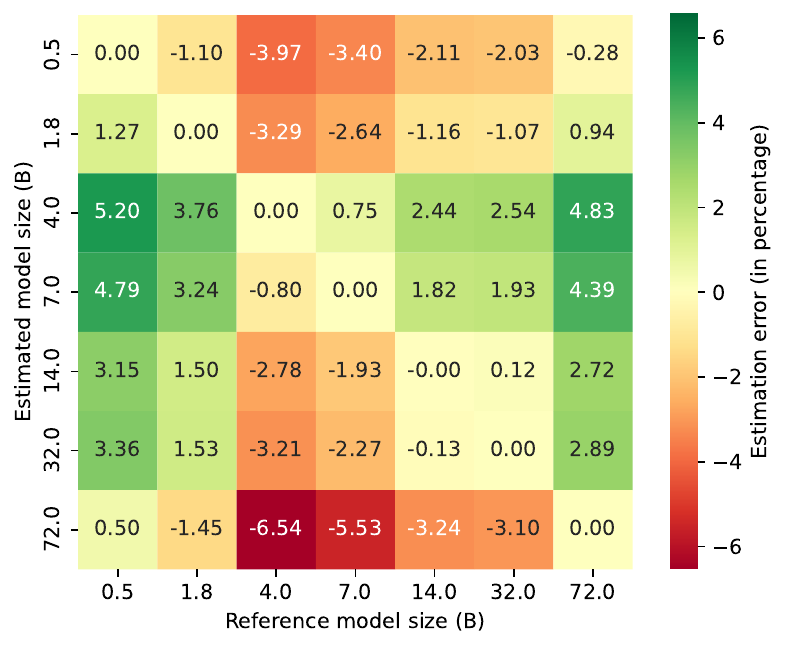}}
    \caption{\textbf{NLL prediction performance based on a single model of a different size within the series.} The value in each cell denotes the relative estimation error (in percentage) with respect to the true tested results. The rows correspond to the sizes of the models being estimated, and the columns represent the sizes of the reference models. The estimation errors remain acceptable despite using only one reference model.}
    \label{fig:nll_prediction}
\end{figure}

Figure~\ref{fig:nll_prediction} presents the NLL prediction errors on the FinePDFs-en dataset for two model series: Qwen3 (from 0.6B to 14B) \citep{yang2025qwen3technicalreport} and Qwen1.5 (from 0.5B to 72B) \citep{bai2023qwentechnicalreport}.
Each heatmap reports the relative estimation error of the predicted NLL when using a single reference model of a given size to estimate the performance of other models within the same series.
This single-reference approach yields highly accurate performance estimates, despite significant disparities between the reference and target model sizes. 
For the Qwen3 series shown in Figure~\ref{fig:nll_prediction}(a), the estimation errors are all tightly bounded within a range of $\pm 3\%$.
Similarly, the Qwen1.5 series in Figure~\ref{fig:nll_prediction}(b) exhibits robust prediction capabilities across a vast parameter span.
For instance, predicting the performance of the 72B model using only the 0.5B model as a reference results in a remarkably low error of 0.50\%.
Overall, these results show that accurate cross-scale NLL prediction can be achieved using only a single evaluated reference model, offering a computationally efficient alternative to conventional scaling-law fitting approaches that require extensive multi-scale training and evaluation.

\begin{figure}[!htb]
    \centering
    \subfloat[Information capacity]{\includegraphics[width=.5\linewidth]{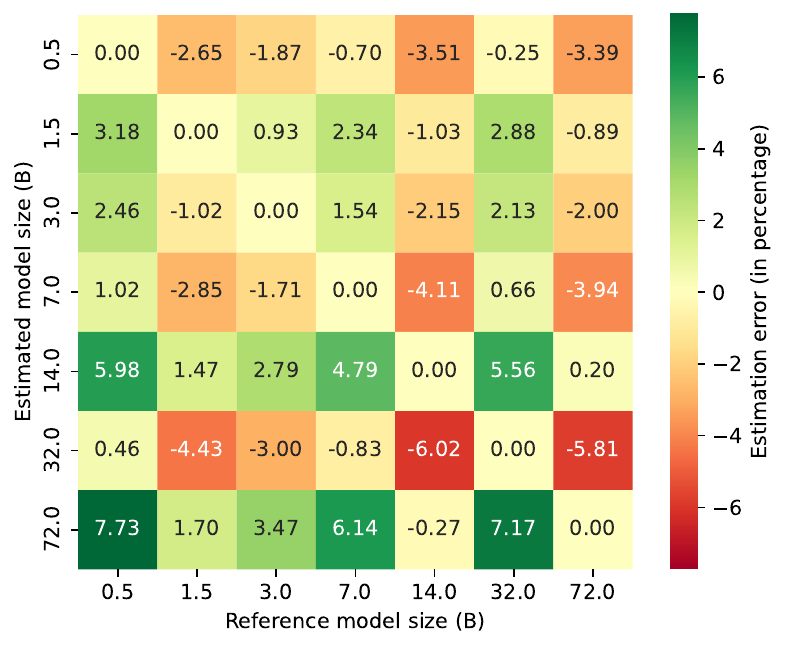}}
    \subfloat[Power law \citep{power_scaling_law}]{\includegraphics[width=.5\linewidth]{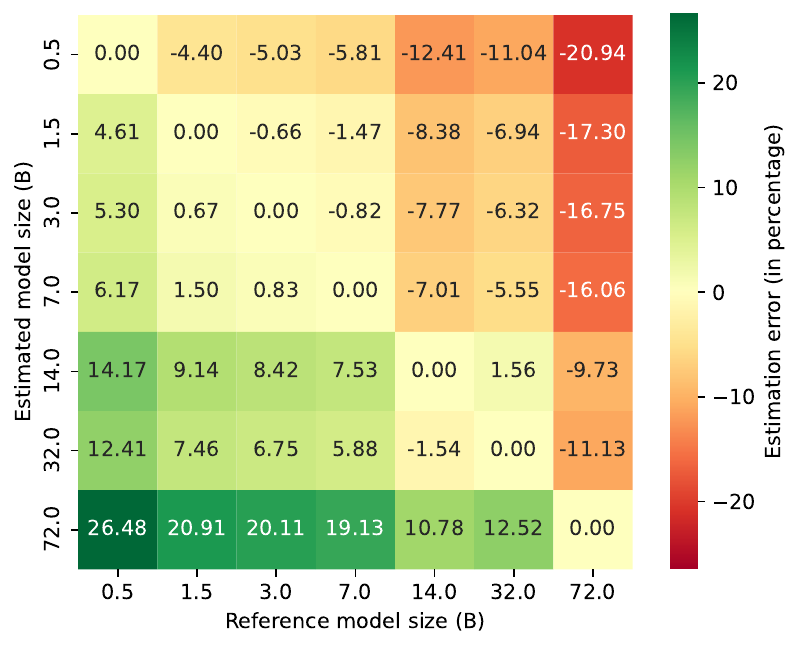}}
    \caption{\textbf{NLL prediction performance of different methods.} The value in each cell denotes the relative estimation error (in percentage) with respect to the true tested result. The rows correspond to the sizes of the model being estimated, and the columns represent the sizes of the reference model. The estimation errors are significantly smaller when based on consistent information capacity in (\ref{eq:NLL_prediction}), compared to the power law in (\ref{eq:NLL_prediction_power_law}).}
    \label{fig:nll_prediction_comparison}
\end{figure}

To demonstrate the advantage of our performance estimation method, we compare the NLL prediction errors based on information capacity with those based on the widely accepted power law \citep{power_scaling_law}.
As one of the few scaling laws that can predict model performance across scales using only a single reference model, the power law is formulated as:
\begin{equation}
    \text{NLL}_{\text{target}} = \text{NLL}_{\text{ref}} \cdot (\frac{ N_{\text{target}}}{N_{\text{ref}}})^{-\alpha_{N}} ,
    \label{eq:NLL_prediction_power_law}
\end{equation}
where the coefficient value $\alpha_{N} = 0.076$ is obtained from the original paper \citep{power_scaling_law}.
Figure~\ref{fig:nll_prediction_comparison} shows the NLL prediction performance of different methods on the FineWeb-Edu dataset for Qwen2.5 series (from 0.5B to 72B) \citep{qwen2025qwen25technicalreport}.
The power law \citep{power_scaling_law} provides a significantly biased estimation, where the estimation error may exceed 25\%.
Predicting the NLL performance of a larger model using a smaller reference model generally results in a positively biased estimation, while the reverse prediction heavily underestimates the NLL.
On the contrary, the information capacity-based method exhibits significantly higher accuracy and stability, with the prediction errors distributed between -6.02\% and 7.73\%.
These empirical results validate that the assumption of consistent information capacity serves as a more robust basis for performance prediction across model scales than the conventional power law.

\subsection{Relation to Benchmark Scores} \label{sec:ic_and_bench}

In this subsection, we provide additional empirical evidence on the relationship between information capacity and benchmark scores.
Models with 7B to 12B activated parameters are selected for comparison, as identical parameter count and inference FLOPs are not possible to maintain among open-source models.
We adopt the highest evaluation results from the official technical reports.

\begin{figure}[!htb]
    \centering
    \captionsetup[subfloat]{labelformat=empty}
    \subfloat{\includegraphics[width=.333\linewidth]{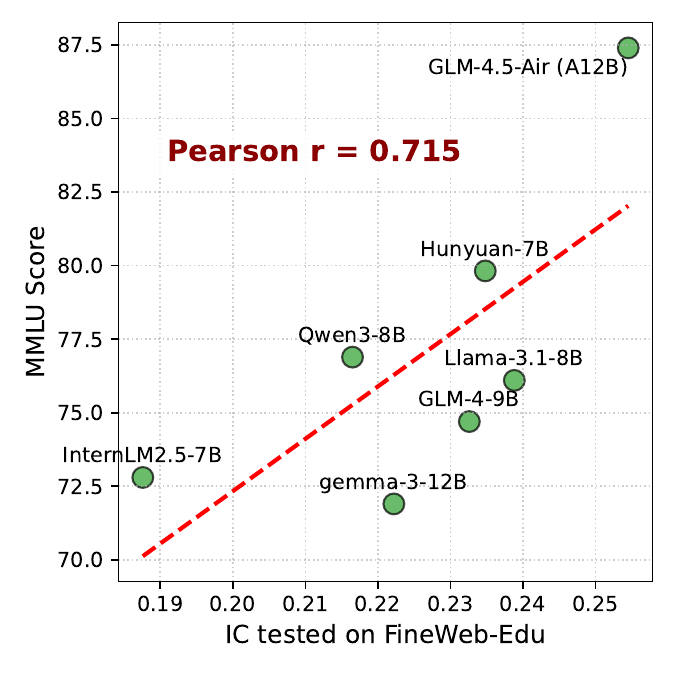}}
    \subfloat{\includegraphics[width=.333\linewidth]{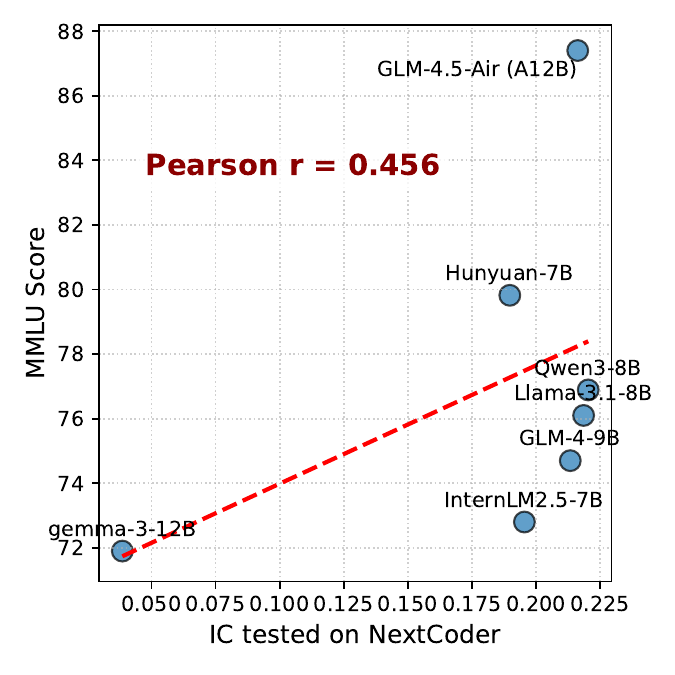}}
    \subfloat{\includegraphics[width=.333\linewidth]{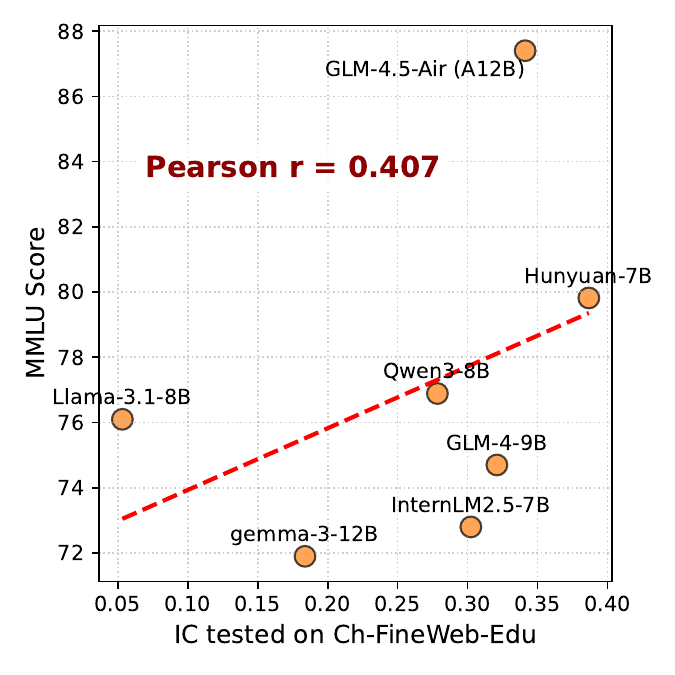}}
    \caption{\textbf{Correlation between MMLU score and information capacity tested on different datasets.} FineWeb-Edu, NextCoder, and Ch-FineWeb-Edu represent typical English, source code, and Chinese corpus, respectively. When the dataset for evaluating information capacity is aligned with the benchmark, benchmark scores and information capacity are most strongly correlated. (IC: Information capacity.)}
    \label{fig:ic_and_mmlu}
\end{figure}

Figure~\ref{fig:ic_and_mmlu} demonstrates the correlation between the massive multitask language understanding (MMLU) score \citep{mmlu} and information capacity tested on FineWeb-Edu, NextCoder, and Ch-FineWeb-Edu datasets, which represent typical English, source code, and Chinese corpus, respectively.
The MMLU benchmark is formulated in English and heavily focuses on Western culture in domains such as history, art, and law.
As a result, the correlation is much stronger when information capacity is tested on the English corpus.
A similar correlation is also observed in coding and Chinese domains.
For instance, the gemma-3 series only achieves an information capacity of 0.0386 on the NextCoder dataset, while the information capacities of Hunyuan, Qwen3, and Ministral-3 series all exceed 0.15.
Correspondingly, gemma-3-12B performs poorly on LiveCodeBench \citep{livecodebench} with a score of 25.7, whereas Hunyuan-7B, Qwen3-8B, and Ministral-3-8B achieve 57.0, 57.5, and 61.6 scores on LiveCodeBench, respectively.
In the C-Eval \citep{c-eval} benchmark consisting of questions from Chinese standardized examinations, Llama-3.1-8B, gemma-3-12B, and Qwen3-8B score 52.0, 61.1, and 83.4, respectively, which are aligned with their information capacities tested on Ch-FineWeb-Edu of 0.0533, 0.1838, and 0.2784.
These results confirm that information capacity can reflect an LLM's capabilities on downstream tasks related to the corpus used for evaluation, and validate our claim in Section~\ref{sec:main_results} that mainstream LLMs deliver imbalanced performance across different domains.

\section{Discussion}

\subsection{Holistic Evaluation}
Information capacity effectively aggregates three aspects in evaluating an LLM's inference efficiency: tokenizer efficiency, task performance, and computational costs per token.
The tokenizer efficiency is represented by the average text size per token, namely the first item of the numerator in (\ref{eq:IC_compute}).
A more efficient tokenizer uses fewer tokens to represent a given input or output text, thus reducing the total FLOPs for an inference request.
The task performance is reflected in the average NLL predicted by the LLM, the second item of the numerator in (\ref{eq:IC_compute}).
This value corresponds to not only the text data size after lossless compression but also the cross-entropy loss during LLM pre-training.
The average FLOPs per token in a logarithmic scale constitutes the denominator in (\ref{eq:IC_compute}), a direct index of computational complexity.
This item is influenced by many aspects of network architecture, including hyperparameter settings, attention mechanism, MoE design, and the structure of feed-forward networks.
Information capacity incorporates these factors into a holistic metric, thus serving as an accurate measure of inference efficiency.

\subsection{Accurate Complexity Measurement}
The central goal of information capacity is to accurately quantify model efficiency across diverse LLM architectures, which necessitates an accurate measurement of model complexity.
Previous scaling laws \citep{power_scaling_law,chinchilla_scaling_law,tian2025greaterleveragescalinglaws} utilize the parameter count as a proxy of inference complexity, which is only viable when all models share a uniform network structure.
For instance, the inference costs vary for LLMs with identical parameter count but different attention mechanisms.
The densing law \citep{xiao2024densinglawllms} attempts to normalize inference complexity with an equivalent parameter count, derived from inverse fitting benchmark scores against a reference model.
However, there is an inherent gap in architecture and training data between the reference model and the model under evaluation, particularly for MoE models with a vastly different architecture.
This disparity introduces potential biases, as the equivalent parameters may not accurately reflect model efficiency.
In contrast, information capacity is directly anchored on FLOPs, a widely accepted metric of computational complexity, thus providing a fair metric robust to variations in model size and architecture.

\subsection{Data Diversity}
Previous scaling laws \citep{power_scaling_law,chinchilla_scaling_law,chen-etal-2025-revisiting} typically employ a single mixed dataset extracted from multiple sources, predominantly consisting of an English corpus.
However, Sections~\ref{sec:main_results} and~\ref{sec:ic_and_bench} have demonstrated the imbalanced performance of LLMs on diverse corpora and benchmarks, highlighting the importance of a comprehensive evaluation.
The densing law \citep{xiao2024densinglawllms} incorporates five benchmarks when measuring the capability density of LLMs, but only reports the maximum density, while the results on each specific benchmark are unavailable.
In this work, we separately present evaluation results on five heterogeneous datasets and compare those results to show the existence of a significant performance bias.

\section{Conclusion}
This paper introduces information capacity, a unified metric of LLM efficiency based on text compression performance relative to computational complexity.
The rationale behind this metric is the correlation between compression and intelligence, as confirmed by the aligned training objective and the empirical evidence from previous studies.
Different from existing metrics, information capacity considers tokenizer efficiency, which affects inference costs but is often neglected in LLM evaluations.
We assess the information capacity of 56 models on 5 heterogeneous datasets and observe a consistent information capacity within a model series and strong linguistic biases in mainstream LLMs. 
Tokenizer efficiency, pretraining data, and the MoE architecture are established as three major factors of information capacity.
Information capacity enables a fair efficiency comparison across model series and accurate performance prediction within a model series.
Given the soaring resource consumption of LLM inference, we anticipate information capacity to be a valuable metric of model efficiency, as opposed to traditional benchmarks on model intelligence.

\bibliographystyle{plainnat}
\bibliography{paper}

\end{document}